\documentclass[a4paper, 10pt, conference]{ieeeconf}      
\usepackage{FG2020}

\FGfinalcopy 

\IEEEoverridecommandlockouts                              
\usepackage{amsmath} 
\usepackage{graphicx}
\usepackage{makecell}
\usepackage{subfig}
\def\rot{\rotatebox}
\def\FGPaperID{11} 

\title{\LARGE \bf
Unique Class Group Based Multi-Label Balancing Optimizer \\ for Action Unit Detection
}

\author{\parbox{16cm}{\centering
		{\large Ines Rieger$^1$$^,$$^*$, Jaspar Pahl\thanks{$^*$These authors contributed equally to this paper.}$^1$$^,$$^*$, Dominik Seuss$^1$}\\
		{\normalsize
			$^1$ Fraunhofer Institute for Integrated Circuits IIS, Erlangen, Germany\\
			\{ines.rieger, jaspar.pahl, dominik.seuss\}@iis.fraunhofer.de}}
	\thanks{This project is funded by the Federal Ministry of Education and Research, grant no. 01IS18056A (TraMeExCo), and the German Research Foundation (PainFaceAnalyzer).}
}

\begin{document}

\ifFGfinal
\thispagestyle{empty}
\pagestyle{empty}
\else
\author{Anonymous FG2020 submission\\ Paper ID \FGPaperID \\}
\pagestyle{plain}
\fi
\maketitle


\begin{abstract}

Balancing methods for single-label data cannot be applied to multi-label problems as they would also resample the samples with high occurrences. We propose to reformulate this problem as an optimization problem in order to balance multi-label data. We apply this balancing algorithm to training datasets for detecting isolated facial movements, so-called Action Units. Several Action Units can describe combined emotions or physical states such as pain. As datasets in this area are limited and mostly imbalanced, we show how optimized balancing and then augmentation can improve Action Unit detection. At the IEEE Conference on Face and Gesture Recognition 2020, we ranked third in the Affective Behavior Analysis in-the-wild (ABAW) challenge for the Action Unit detection task.

\end{abstract}


\section{INTRODUCTION}

Imbalanced datasets present a major challenge when training models for a multi-label multi-class problem such as facial expression recognition, text or music label categorization. Facial expression datasets often contain underrepresented classes, since different expressions do not occur naturally with the same frequency. But in contrast to single-label data, resampling methods such as simple over- or undersampling that clone over- or underrepresented samples with a majority or minority label do not work, as they would also increase or decrease potential other labels in the same sample as well.

Facial expressions such as physical pain \cite{kunz_facial_2019} or the basic emotions anger, disgust, fear, happiness, sadness, and surprise \cite{ekman2003darwin} can be described by Action Units.
Action Units (AUs) are distinct facial movements, which are described in the Facial Action Coding System (FACS) \cite{ekman2002facial}. There are several approaches to automatically detect the occurrence and intensity of AUs \cite{martinez2017automatic, rieger2020Verifying}. Deep learning models are one of the most successful approaches, but require large amounts of data to perform well. But the manual annotation time for AU occurrences and intensity is high and must be conducted by expert FACS-coders. Thus, the training data for AUs is limited. Furthermore, AU datasets are mostly imbalanced as different AUs do not occur with the same frequency for different facial expressions.

We propose a fast and simple, yet effective balancing and augmenting approach framed as optimization problem to compensate for imbalanced and scarce occurrences in a multi-label dataset. We apply this algorithm on the use case of Action Unit detection and show that our balancing approach can enhance the detection of AUs.

\begin{figure*}[ht!]
	\centering
	\subfloat[Our balancing pipeline based on unique class label groups ]{{\includegraphics[width=6cm]{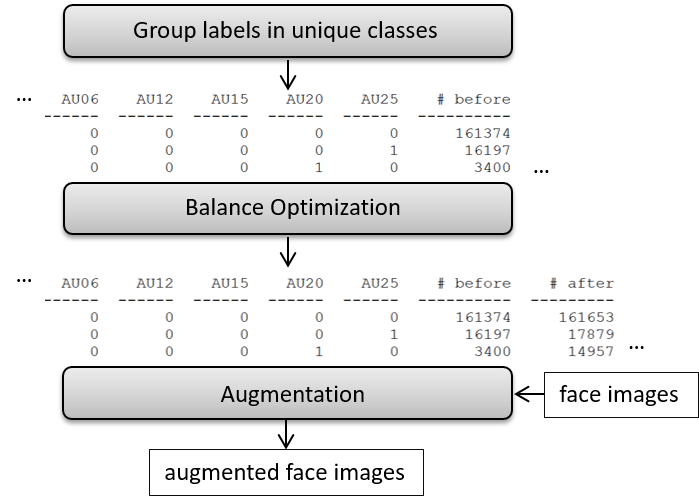} }}%
	\subfloat[ResNet18-112 with 18 layers and input images of size 112x112 pixels, adapted from \cite{rieger2019towards}]{{\includegraphics[width=12cm]{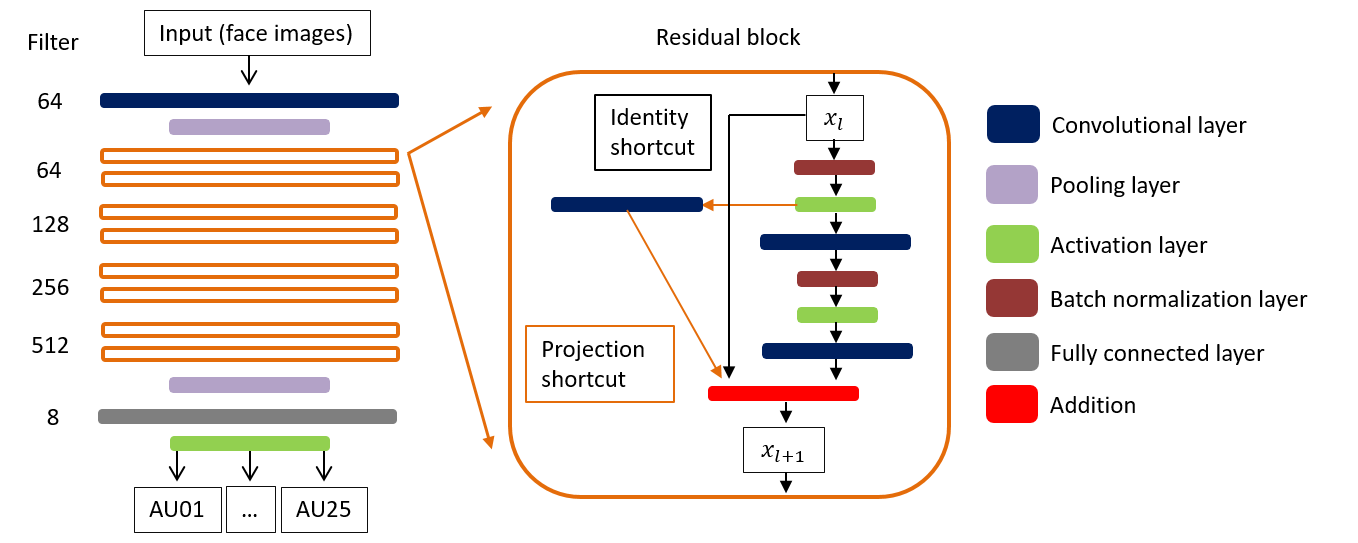} }}%
	\caption{This shows our (a) balancing pipeline and (b) training backbone}
	\label{pipeline}
\end{figure*}


\section{RELATED WORKS\label{related}}

\subsection{Balancing Multi-Label Data}
Single-label resample techniques are often extended to be also used for multi-label data \cite{zhang2013review}.
Charte et al. \cite{charte2015addressing} propose the LP-RUS and LP-ROS algorithms for random under- and oversampling. As in our approach, they first transform the multi-label dataset in a multi-class dataset, handling each distinct combination of labels as one labelset class. They first determine minority and majority labelsets, which include a label belonging to an under- or over represented class. Then, these classes are over- or undersampled by cloning or removing samples until they reach the average mean size of the labelset classes. Charte et al. \cite{charte2015addressing} further propose the ML-RUS and ML-ROS algorithms for under- and oversampling, where they evaluate the individual imbalance level of each sample and thus blocking certain samples to be removed or cloned. 

MLSMOTE (Multilabel Synthetic Minority Over-sampling Technique) \cite{charte2015mlsmote} is an extension of the SMOTE \cite{chawla2002smote} algorithm in order to oversample minority classes. The SMOTE algorithm creates new samples by taking a new instance among the nearest neighbors of the desired underrepresented sample. As Charte et al. \cite{charte2015addressing}, they determine minority labelset classes. Charte et al. \cite{charte2015mlsmote} create new samples for these by using the label correlation information. By experiments, they show that MLSMOTE outperforms the ML-ROS algorithm.

In contrast to these approaches, we formulate the balancing of datasets as an optimization problem and then use slight augmentation techniques to increase the number of samples.

\subsection{Action Unit Detection}
The following approaches also use the AffWild2 validation and testing dataset for evaluation (see Section \ref{results}). Kollias et al. \cite{kollias2020analysing} use a MobileNetV2 \cite{sandler2018mobilenetv2} as backbone for Action Unit Detection. Deng et al. \cite{deng2020fau} propose a multitask model for Valence and Arousal, Emotion and Action Unit Detection. They use partial labels in a teacher and student training setting. For tackling the imbalanced training dataset, Deng et al. \cite{deng2020fau} employ the ML-ROS algorithm \cite{charte2015addressing}. Kuhnke et al. \cite{kuhnke2020two} propose a two-stream multitask model for detecting Valence and Arousal, Emotions and Action Units. They use features from an aural and a visual stream employing spatial and temporal convolutions.


\section{DATASETS\label{datasets}}

We use the Aff-Wild2 \cite{kollias2018aff, kollias2019expression,kollias2018multi, kollias2019deep} as training, validation and test dataset and as additional training datasets the Emotionet \cite{emotionet}, the Actor Study \cite{seuss2019emotion} and the Extended Cohn-Kanade (CK+) \cite{kanade2000comprehensive,lucey2010extended}.

The Aff-Wild2 database is the largest in-the-wild database consisting of videos and is partly annotated for the three tasks valence-arousal estimation \cite{zafeiriou2017aff}, basic expression recognition \cite{kollias2017recognition} and Action Unit detection. All videos were scraped from YouTube. For Action Unit detection, 56 videos with  63  subjects  (32  males and  31  females) were annotated by experts. In total, 398,835 frames are annotated. For the challenge, this dataset is divided in training, validation and testing. This dataset is for non-commercial research purposes only.

The Emotionet database consists of approximately one million images scraped from the web using emotive keywords. Benitez-Quiroz et al. \cite{fabian2016emotionet} analyzed the occurrence and intensity of AUs in 90\% of the images automatically. Experts FACS-coded 10\% of the dataset manually. We only use the manually coded part of the dataset for our experiments. This dataset is for non-commercial research purposes only.

The Actor Study contains sequences of 21 actors, in total 68 minutes of video filmed from different views and camera speed. Each actor had the task to display specific Action Units in different intensities and had to respond to scenarios and enactments. The frames are annotated with Action Units and their corresponding intensities. We use the center view of the low speed camera for our approach. This dataset was recently published and will be made publicly available for commercial research.

The CK+ dataset contains 593 videos of 123 subjects. The sequences show the facial expressions from neutral to strong. All expressions are posed. The presence or absence of the Action Units is coded for the peak frames, the strongest expression, only. This dataset is for non-commercial research purposes only.


\section{METHODS\label{methods}}

\subsection{Pre-processing\label{Preprocessing}}
For the AffWild2 training, validation, and testing dataset we use the cropped and aligned images of 112 x 112 pixel size provided by the ABAW 2020 challenge in order to eliminate sources of error and make our approach easier to compare with others. Please note that there is noise in this precropped dataset, as the main person is not identified at all frames correctly and thus, the annotated labels do not match the facial image. The remaining training sets had no pre-cropped images available. For the CK+ dataset and the Actor Study dataset the Sophisticated High-speed Object Recognition Engine SHORE\texttrademark~\cite{kublbeck2006face} is used to crop images of the same size. The Emotionet dataset is cropped using OpenFace~\cite{amos2016openface}. All images were converted to gray scale and pixel values were normalized in the range of $[0,1]$.

\subsection{Balancing\label{Balancing}}
Combining all datasets from Section~\ref{datasets} yields a highly imbalanced dataset (see "non-aug" in Table~\ref{tab_aus}). Imbalanced datasets cause problems in the detection of under-represented Action Units. Therefore, we implement a class balancing algorithm (Fig. \ref{pipeline}a) which increases the allowed number of augmentations in the under-represented Action Units. 
In a multi-label multi-class problem, the occurrences of a class cannot be increased by simply augmenting a single image multiple times because the potential other classes in the image would be increased by the same amount. Therefore, we formulate the class balancing as an optimization problem:

\begin{subequations}
	\begin{equation}
	f(n\_u) = \sum \mid z - \sum^{N\_u}_{i} \frac{z_{i}}{N\_u}  \mid + \lambda \ Var \left( \frac{n\_u}{n\_o}\right)
	\end{equation}
	\begin{equation}
	z = \sum_{r}  n\_u_{r} * y\_u_{r},
	\end{equation}
\end{subequations}

\begin{figure}[ht]
	\centering
	\includegraphics[scale=0.57]{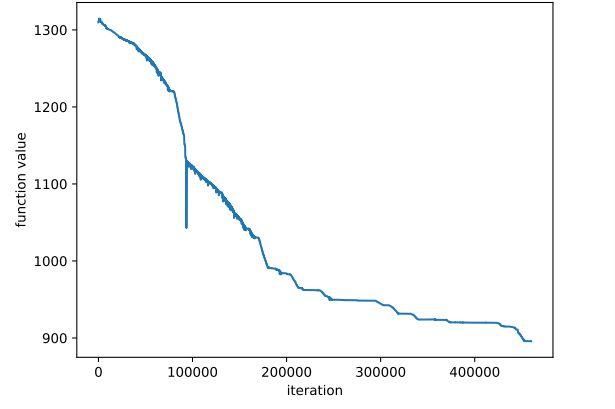}
	\caption{Convergence of balance optimization with  $\lambda = 70$ and $n\_o < n\_u < 10 \ n\_o$}
	\label{optimizer}
\end{figure}

where $n\_u$ is the vector of number of occurrences of the images with a unique Action Unit combination and $y\_u$ is the respective one-hot-encoded array of labels for the unique labelsets. The operator $*$ between $n\_u_{r}$ and $y\_u_{r}$ denotes a scalar multiplication for each row $r$. The variable $n\_o$ is the vector with the original occurrences in the dataset non-aug, and $\lambda$ is a weighting parameter. We used $\lambda = 70$ as weighting value for our experiments (Fig. \ref{optimizer}). Furthermore, we restricted the search domain for $n\_u$ as:
\begin{equation}
n\_o < n\_u < 10 \ n\_o.
\end{equation}
As such, we can control the number of additional augmentations while stopping the optimizer from increasing the relative growth of a single unique label combination too much. The latter would cause problems due to too many augmentations on a single image, which potentially leads to overfitting.

We go through the images and augment them according to the optimized occurrences in their respective labelset class until the max occurrence is reached. The augmentations are implemented using the imgaug library~\cite{imgaug}. We use Flipping\_lr, Gaussian Blur, Linear Contrast, Additive Gaussian Noise, Multiply, and Perspective Transform. These slight augmentation methods are suited for Action Units, as the labels are still valid after the augmentation.

Table~\ref{tab_aus} shows the label distribution of the merged training datasets, the distribution of the total training dataset "non-aug" and the augmented dataset "aug-1", augmented by the upper algorithm. We can see that the algorithm is working well, because Action Units like the AU01 with a high occurrence in the non-aug dataset are very little augmented, while Action Units like the AU16, with a low occurrence in the non-aug dataset are now significantly better represented.

\begin{table}[htb!]
	\centering
	\caption{Label distribution of the original training datasets and of the augmented dataset	\label{tab_aus}}
	\begin{tabular}{r||r|r|r|r||r|r}
		\rot{90}{Dataset} & \rot{90}{AffWild} & \rot{90}{\makecell{10\%\\Emotionet}} &\rot{90}{CK+}  & \rot{90}{Actorstudy} & \rot{90}{\makecell{total train set \\ non-aug}} & \rot{90}{\makecell{total augmented \\ train set aug-1}} \\
		\hline 
		\hline 
		\rule{0pt}{10pt}
		AU01 & 47548 & 589 & 2964 & 11079 & 62180 & 62874 \\ 
		\hline 
		\rule{0pt}{10pt}
		AU02 & 2271 & 0 & 1875 & 8553 & 12699 & 24754 \\
		\hline 
		\rule{0pt}{10pt}
		AU04 & 32387 & 676 & 3753 & 11393 & 48209 & 48365 \\ 
		\hline 
		\rule{0pt}{10pt}
		AU06 & 9290 & 0 & 2174 & 6991 & 18455 & 33091 \\ 
		\hline 
		\rule{0pt}{10pt}
		AU12 & 22964 & 0 & 2475 & 7798 & 33237 & 35417 \\
		\hline
		\rule{0pt}{10pt} 
		AU15 & 1537 & 0 & 1654 & 2115 & 5306 & 17748 \\
		\hline 
		\rule{0pt}{10pt}
		AU20 & 3490 & 0 & 1406 & 4704 & 9600 & 25681 \\
		\hline 
		\rule{0pt}{10pt}
		AU25 & 7463 & 1780 & 5359 & 13573 & 28175 & 35416 \\ 
		\hline 
		\hline
		\rule{0pt}{10pt}
		\makecell{total \\ images}  & 232842 & 20160 & 10724 & 58825 & 322551 & 378041 \\ 
	\end{tabular} 
\end{table}

\subsection{Evaluation Metric\label{metric}}
The results are measured in F1 score (\ref{metric}a) for each Action Unit. Furthermore, the results are measured according to the ABAW Challenge metric \cite{kollias2020analysing}, denoted here as AWC (\ref{metric}c). AWC is the unweighted average of the accuracy Acc (\ref{metric}b) and the F1 macro score. The variable $N$ stands for all predictions.

\begin{subequations}
	\begin{equation}
	F1 = \frac{2 \cdot precision\cdot recall}{precision+ recall}
	\end{equation}
	\label{metric}
	\begin{equation}
	Acc = \frac{N\_correct\_predictions}{N}
	\end{equation}
	\begin{equation}
	AWC = \frac{1}{2}\left(\frac{\sum^{N\_AU}_{i} F1_i}{N\_AU} + Acc\right)
	\end{equation}
\end{subequations}

\subsection{Training}
Our backbone is a parameter-reduced 18-layer ResNet with input images of size 112x112 pixels (ResNet18-112) \cite{rieger2019towards} (Fig. \ref{pipeline}b). This modified ResNet is based on He et al. \cite{he2016deep}. We use the ReLU activation function and changed the output layer to a sigmoid activation function with eight outputs according to the number of Action Units. The threshold for a positive detection is 0.5. We use a weighted F1 loss function with an Adam optimizer and a learning rate of 0.0001. These parameters have been determined by grid search.


\section{RESULTS\label{results}}

Table \ref{tab_results_aus} shows our results on the Aff-Wild2 validation dataset. The results show that our balanced augmentation algorithm can enhance the results of a multi-label problem like Action Unit Detection. Our model trained on the augmented dataset aug-1 performs better at detecting all Action Units equally, as classes are now detected that could not be detected before and the overall F1 macro score increases from 0.21 to 0.24.

Our trained models supersede by far the baseline model of Kollias et al. \cite{kollias2020analysing}, who are also using a single-task model. Our model trained on the augmented dataset aug-1 performs slightly worse in comparison to the multitask models of Deng et al. \cite{deng2020fau} and Kuhnke et al. \cite{kuhnke2020two}, who were ranked first and second in the ABAW 2020 challenge. In contrast to Deng et al. \cite{deng2020fau} and Kuhnke et al. \cite{kuhnke2020two}, we use the noisy precropped AffWild2 dataset, which has a negative effect on the performance. But what is apparent is that our model trained on aug-1 generalizes better on the testing data than the model trained on the original dataset non-aug. There is 0.5 difference between the F1 macro scores of the validation and testing dataset of the model trained on the non-aug dataset and 0.2 difference of the model trained on the aug-1 dataset. 

\begin{table}[htb!]
	\centering
	\caption{Our results on the Aff-Wild2 validation dataset. The results of the AUs are measured in F1 Score. Best results are in bold.\label{tab_results_aus}}
	\begin{tabular}{c||c|c}
		Model & ResNet18-112 (ours) & ResNet18-112 (ours) \\ 
		\hline
		Train Set & non-aug & aug-1 \\ 
		\hline 
		\hline 
		\rule{0pt}{10pt}
		AU01  & 0.64 & \textbf{0.71}\\
		\hline
		\rule{0pt}{10pt}
		AU02 & 0.00 & 0.00\\
		\hline
		\rule{0pt}{10pt}
		AU04  &\textbf{0.47} & 0.36\\
		\hline
		\rule{0pt}{10pt}
		AU06  & 0.21 & \textbf{0.33}\\
		\hline
		\rule{0pt}{10pt}
		AU12  & 0.39 & \textbf{0.41}\\
		\hline
		\rule{0pt}{10pt}
		AU15  &0.00 & \textbf{0.10}\\
		\hline
		\rule{0pt}{10pt}
		AU20  & 0.00 & \textbf{0.04}\\
		\hline
		\rule{0pt}{10pt}
		AU25  &0.00 & 0.00\\
		\hline
		\hline
		\rule{0pt}{10pt}
		F1 macro & 0.21 & \textbf{0.24}\\
		\hline
		\rule{0pt}{10pt}
		Acc & \textbf{0.94} & 0.93\\
		\hline
		\rule{0pt}{10pt}
		AWC & 0.58 & \textbf{0.59}\\
	\end{tabular} 
\end{table}

\begin{table}[htb!]
	\centering
	\caption{Comparable results on the Aff-Wild2 validation and testing dataset. Best results are in bold, second best results are in brackets.	
		\label{tab_results_aus}}
	\begin{tabular}{c||c|c||c |c|c}
		& F1Macro & AWC & F1Macro & Accuracy & AWC \\ 
		\hline
		\rule{0pt}{10pt}
		Dataset & \multicolumn{2}{|c||}{Validation} & \multicolumn{3}{c}{Testing} \\ 
		\hline 
		\hline 
		\rule{0pt}{10pt}
		baseline \cite{kollias2020analysing} & -&0.31 & -&- & 0.26\\
		\hline 
		\rule{0pt}{10pt}
		twostream \cite{kuhnke2020two} & - & [0.59] & [0.27] & \textbf{0.93} & [0.60] \\
		\hline 
		\rule{0pt}{10pt}
		multitask \cite{deng2020fau} & - &\textbf{ 0.63} &\textbf{ 0.31}& [0.91]& \textbf{0.61}\\
		\hline 
		\rule{0pt}{10pt}
		non-aug (ours) & 0.21 & 0.58& 0.16 & [0.91]& 0.53 \\
		\hline 
		\rule{0pt}{10pt}
		aug-1 (ours) & 0.24 & [0.59]& 0.22 & 0.89 & 0.55 \\
	\end{tabular} 
\end{table}

\section{CONCLUSIONS AND FUTURE WORKS}

In this paper we present an approach based on a multi-label class balancing algorithm as a pre-processing step to overcome the imbalanced occurrences of Action Units in the training dataset. We train a ResNet with 18 layers and show an improvement in detecting single AUs better and generalizing better on unseen data when using the augmented training dataset in contrast to the original training dataset. For future work, we plan on enriching our approach with a multi-task framework in order to improve our results. Furthermore, we plan on comparing our results on the aug-1 dataset to results on even further augmented datasets created with a higher augmentation rate, i.e. different weightings and compare our results thoroughly with other resampling algorithms.

\bibliographystyle{plain}
\bibliography{bib}

%
%
%
%

\end{document}